\documentclass[conference]{IEEEtran}
\IEEEoverridecommandlockouts
\usepackage{cite}
\usepackage{amsmath,amssymb,amsfonts}
\usepackage{graphicx}
\usepackage{textcomp}
\usepackage{xcolor}
\usepackage{url}
\usepackage{hyperref}

\usepackage{amsthm} 
\usepackage{algorithm}       
\usepackage{algpseudocode}   

\theoremstyle{plain}
\newtheorem{lemma}{Lemma}

\def\BibTeX{{\rm B\kern-.05em{\sc i\kern-.025em b}\kern-.08em
    T\kern-.1667em\lower.7ex\hbox{E}\kern-.125emX}}
\begin{document}

\title{Deep Transductive Outlier Detection}

\author{\IEEEauthorblockN{Simon Klüttermann}
\IEEEauthorblockA{
\textit{TU Dortmund University}\\
Dortmund, Germany \\
Simon.Kluettermann@cs.tu-dortmund.de}
\and
\IEEEauthorblockN{Emmanuel Müller}
\IEEEauthorblockA{
\textit{TU Dortmund University}\\
Dortmund, Germany \\
emmanuel.mueller@cs.tu-dortmund.de}
}

\maketitle

\begin{abstract}
Outlier detection (OD) is one of the core challenges in machine learning. Transductive learning, which leverages test data during training, has shown promise in related machine learning tasks, yet remains largely unexplored for modern OD. We present \textsc{Doust}, the first end-to-end transductive deep learning algorithm for outlier detection, which explicitly leverages unlabeled test data to boost accuracy. On the comprehensive ADBench benchmark, \textsc{Doust} achieves an average ROC-AUC of $89\%$, outperforming all 21 competitors by roughly $10\%$. Our analysis identifies both the potential and a limitation of transductive OD: while performance gains can be substantial in favorable conditions, very low contamination rates can hinder improvements unless the dataset is sufficiently large.
\end{abstract}


\section{Introduction}

Outlier detection (OD) is critical for a wide range of applications, ranging from fraud detection~\cite{fraudapl,auto_appl_electionfraud} and network intrusion prevention~\cite{auto_appl_httptraffic,auto_appl_networkintrusion} to medical diagnosis~\cite{medapl,medicalad} and industrial fault monitoring~\cite{machinefault,auto_appl_faultsinparticleaccelators}. Its importance continues to grow as the volume and complexity of real-world data increase, making the identification of rare, atypical, or anomalous instances a key requirement for reliable decision-making.

Despite decades of progress, OD has hit a performance plateau: deep models fail to outperform shallow ones: While modern deep neural architectures have been introduced for OD (e.g. \cite{dte,DEAN,dagmm}), empirical evidence suggests that they often lag behind well-established shallow methods~\cite{surveyzhao}, or deliver only statistically insignificant improvements~\cite{dte}. 
This observation motivates the need for a paradigm shift: instead of further refining existing inductive frameworks, we propose to explore a fundamentally different setting.

\begin{figure}[htbp]
    \centerline{\includegraphics[width=1.0\linewidth]{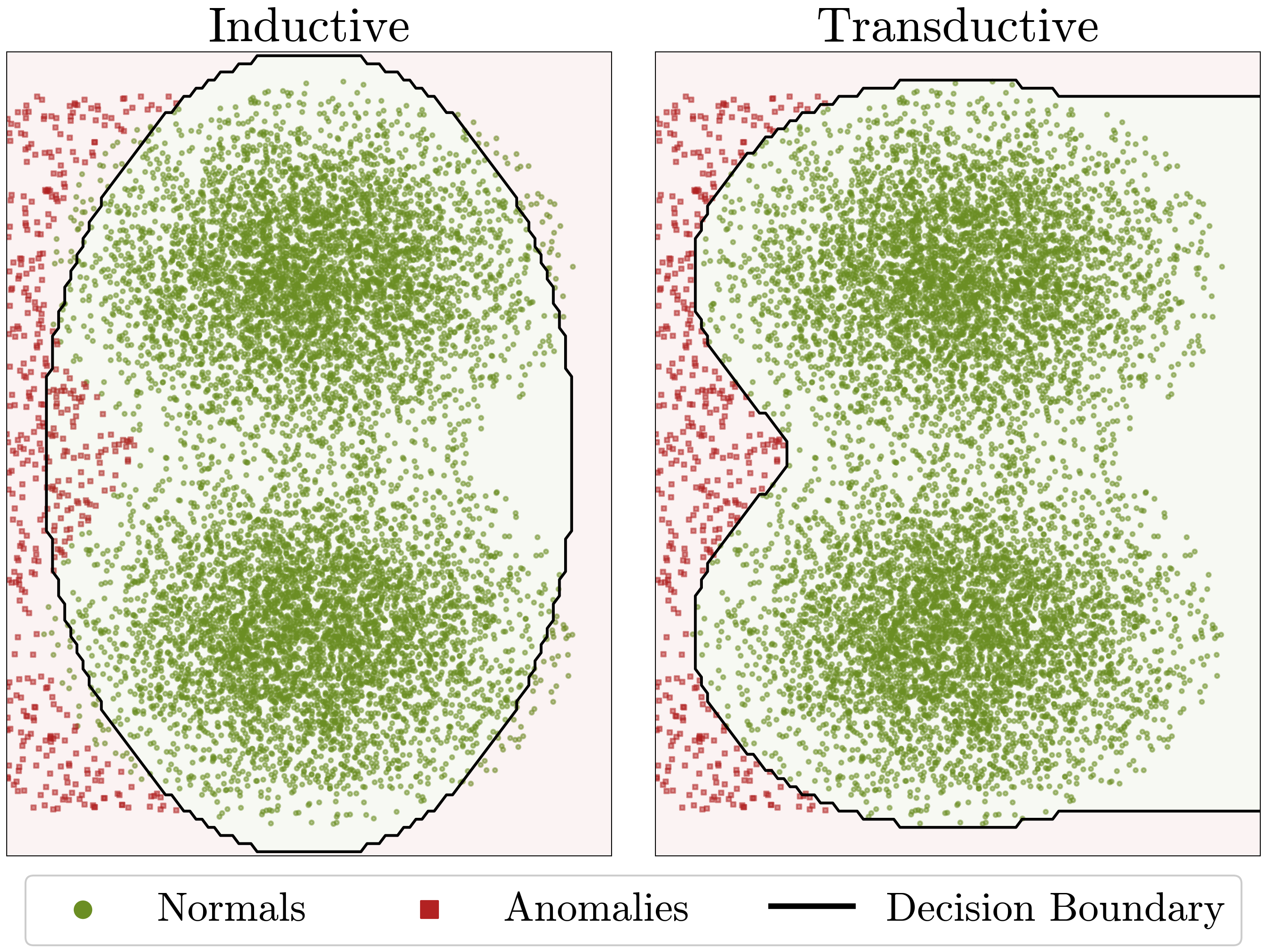}}
    \caption{Toy example of the difference between an inductive and transductive outlier detection algorithm. While the inductive algorithm (left) tries to learn the shape of the normal data, the transductive algorithm (right) can use the test data to learn the decision boundary between normal and abnormal regions much more accurately.}
    \label{fig:toy}
\end{figure}

In this work, we suggest focusing on \emph{transductive} outlier detection. Unlike the common inductive setting, where a model is trained solely on a given training set and subsequently applied to unseen data, the transductive setting allows the model to explicitly incorporate information from the unlabeled test set during training. This distinction is crucial: transductive methods can be tailored to a specific dataset and thus often achieve stronger performance in that specific context, even though they will generalize less well to new datasets. We visualize this difference on a toy dataset in Figure~\ref{fig:toy}. 

Despite its apparent benefits, research on transductive OD remains limited. Recent work has either addressed transduction only implicitly~\cite{boostingAD1,boostingAD,schubert} or designed approaches applicable only to narrow problem scenarios~\cite{lemanTransductive,lemanTransductiveVision}.

We argue that the transductive setting holds particular promise in the most common OD scenario: the one-class setting (sometimes also referred to as unsupervised in the literature~\cite{metasurvey}), where two datasets are available. The first contains only normal instances, while the second (the target or test set) is unlabeled and potentially contaminated with anomalies. Most existing methods adopt an inductive approach, training exclusively on the normal dataset and ignoring the information available in the contaminated test set. Our key insight is that we can explicitly leverage the differences between the training and unlabeled test data to improve performance. Importantly, this approach remains unsupervised and does not violate data leakage rules, as we do not use the test labels.

We make the following contributions:
\begin{itemize}
    \item We introduce the first deep learning method that explicitly incorporates unlabeled test data in OD training, enabling deep transductive outlier detection.
    \item We demonstrate that our approach consistently outperforms $21$ competitive methods across $121$ datasets, including both modern deep learning models and strong shallow baselines, achieving nearly $10\%$ higher average ROC-AUC than the best alternative.
    \item We identify and formalize a previously unrecognized fundamental limitation of transductive OD methods, showing that not all tasks can be improved using transduction, but also that this limitation diminishes as dataset size increases.
    \item To foster reproducibility, we provide our code and individual results at \href{https://anonymous.4open.science/r/doust-15EE/}{https://anonymous.4open.science/r/doust-15EE}
\end{itemize}

\section{Related Work}

\subsection{Outlier Detection}\label{sec:rw:od}
Outlier detection~\cite{hawkinsdef1970fischer} (OD), also referred to as anomaly detection (AD) or novelty detection~\cite{surveyruff}, aims to identify data instances that deviate significantly from the majority of observations~\cite{metasurvey}. A wide variety of approaches have been proposed, ranging from classical statistical methods and distance-based techniques to modern ensemble methods and deep learning models~\cite{charu-book-outlier-analysis}. OD can be studied in various settings, which differ by the amount of labeled data that is available~\cite{surveyzhao}. Since labeling rare anomalies is very expensive, the most common setting is the one-class classification (only some normal samples are known), which is also sometimes referred to as unsupervised or semi-supervised anomaly detection in the literature~\cite{metasurvey}. Here, models have to learn a sense of normality from a given dataset and aim to find any type of anomaly in a second dataset that deviates from this sense of normality. Because this can be a challenging task, many OD algorithms have been proposed over the years. We compare here against $21$ different ones. These range from distance-based methods (KNN~\cite{knn}, LOF~\cite{lof}, SOD~\cite{sod}, CBLOF~\cite{cblof}), over surrogate task algorithms (Autoencoder (AE)~\cite{aean}, DEAN~\cite{DEAN}, SEAN~\cite{sean}, DTE~\cite{dte}, DeepSVDD~\cite{deepsvdd}, OCSVM~\cite{ocsvm}, PCA~\cite{pca}, NeuTral~\cite{NeuTralAD}, GOAD~\cite{goad}) to density estimation approaches (IFOR~\cite{ifor,iforisdensity}, Normalizing Flows (NF)~\cite{nf}, Variational Autoencoder (VAE)~\cite{vae}, HBOS~\cite{hbos}, COPOD~\cite{copod}, ECOD~\cite{ecod}, LODA~\cite{loda}, DAGMM~\cite{dagmm}). Overall, this list includes still strong shallow competitors proposed in 2000~\cite{knn,lof}, and modern deep learning ensembles as proposed in 2025~\cite{DEAN}.

\subsection{Transductive Machine Learning}
Machine learning methods can broadly be categorized into \emph{inductive} and \emph{transductive} paradigms. Inductive methods learn a generalizable function from training data, aiming to perform well on any unseen sample from the same distribution. In contrast, transductive methods optimize directly for the specific test set at hand, allowing them to exploit patterns in the unlabeled test data to improve predictive performance~\cite{vapnik1998transduction}. In fact, Vapnik argued that transduction was often the more natural learning setting, since it avoids the unnecessary step of inferring a general rule prior to making predictions on a fixed evaluation set.

Prominent examples of transductive learning include transductive support vector machines (TSVM)~\cite{joachims1999tsvm}, graph-based learning approaches that operate on a fixed input graph~\cite{zhou2004llgc,zhu2002lp}, and certain non-parametric methods such as $k$-nearest neighbors~\cite{knn}, which often directly combine the training and test set. Transductive learning is often more accurate in practice because it can align its decision boundary with the structure of the actual evaluation set. 

While transduction was studied primarily in the early days of machine learning, it inspired several adjacent paradigms that deal with generalization, such as meta-learning~\cite{metalearning} and domain adaptation~\cite{domainadaptation}. More recently, it has seen a resurgence under the related concept of \emph{test-time training}~\cite{testtimetraining,testtimefollowupAE,wang2021tent}, where models are adapted or fine-tuned using the unlabeled test data prior to inference. While test-time training is frequently explored in supervised or semi-supervised contexts, the core idea (leveraging unlabeled evaluation data to refine a model) extends naturally to other domains, including OD.

\subsection{Transductive Outlier Detection}
OD is an especially suitable domain for transductive learning. The challenge with transductive learning and especially modern test-time-training approaches often is that adapting a trained supervised model generally requires labels. One way to circumvent this is to train a secondary, unlabeled task at the same time. However, since OD methods are already trained with no/limited labels, we can circumvent this problem.

Thus, the field of \emph{transductive outlier detection} dates back to early OD research~\cite{vovk2005cp,michigan2005transductivead}, many unsupervised approaches were effectively transductive~\cite{knn,lof,ifor} in the sense that there was no separation between training and test dataset. However, these early approaches did not fully capitalize on the benefits of transduction. With the availability of more data and the rising prominence of neural networks, the concept of transduction was largely forgotten.

To date, the field is largely unexplored. One notable exception is the work of Prof. Akoglu~\cite{lemanTransductiveVision,lemanTransductive}, which implicitly uses test data for model selection. Another related thread is found in certain ensemble-based approaches, such as boosting frameworks that iteratively adjust base learners using feedback from the evaluation set~\cite{boostingAD,boostingAD1,schubert}. However, explicit formulations of transductive OD and especially dedicated deep learning algorithms that capitalize on this setting remain rare.

Our approach addresses this gap by introducing a deep outlier detection method that leverages unlabeled test data during training, enabling the model to adapt to each test set and realize the performance gains in outlier detection originally envisioned by transductive learning.

\section{Deep Transductive Outlier Selection}

\begin{figure}[htbp]
    \centering
    \setlength{\fboxsep}{0pt} 
    \fbox{\includegraphics[width=1.0\linewidth]{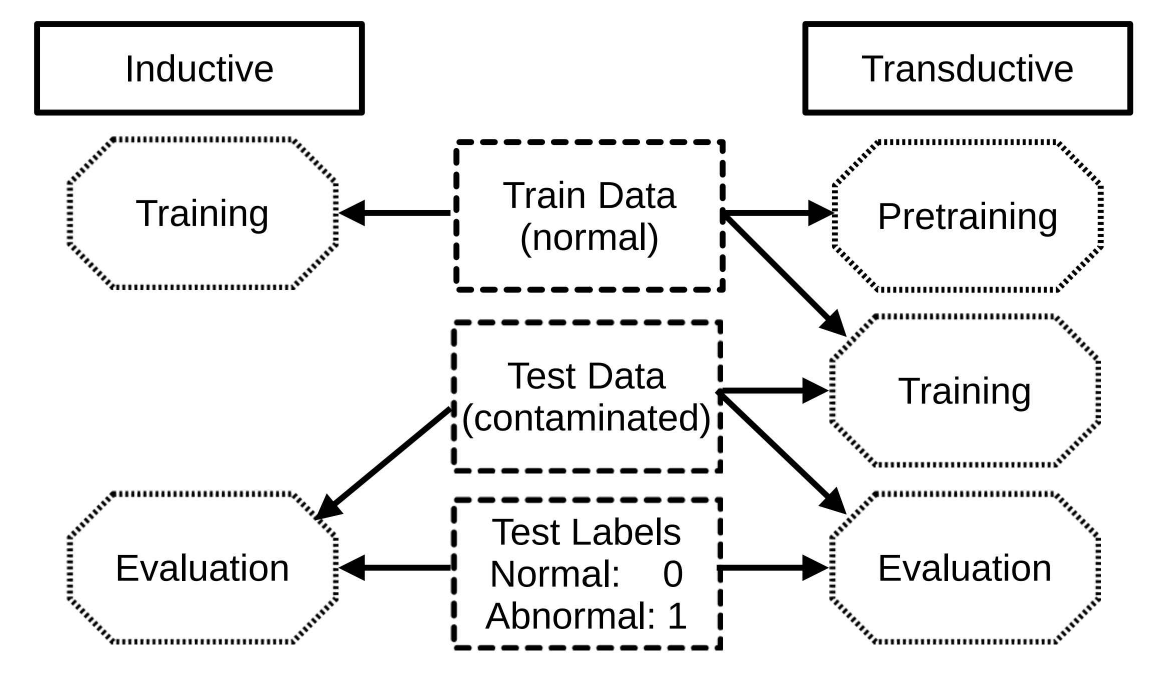}}
    \caption{Comparison of the approach of an inductive outlier detection algorithm (left) to a transductive one (right). While an inductive algorithm only uses the training data for training, a transductive one also uses the test data to specialize the model to this test set. This does not violate the standard separation between training and test data, as the test labels remain unused.}
    \label{fig:point}
\end{figure}

We propose \emph{Deep OUtlier Selection by Transduction} (\textsc{Doust}), a novel deep learning framework for OD that explicitly incorporates unlabeled test data into the training process for OD. \textsc{Doust} builds upon the DEAN framework~\cite{DEAN}, which tries to learn surrogate tasks for OD. Unlike DEAN, \textsc{Doust} introduces a two-stage optimization procedure designed to exploit the structural differences between the purely normal training set and the potentially contaminated test set. Consequently, \textsc{Doust} seeks to learn an optimal surrogate task for a specific OD problem. We show a summary of this training process in Figure~\ref{fig:point}.

\subsection{Model Overview}
As it is common in one-class classification OD, our model operates on two datasets: a training set $X_{\mathrm{train}}$ containing only normal instances, and a test set $X_{\mathrm{test}}$ that may contain both normal and abnormal instances. Our goal is now to train a neural network $f:\mathbb{R}^d \to [0,1]$, where higher scores $f(x)$ correspond to a greater likelihood of $x$ being abnormal.

\paragraph{Stage 0: Baseline Representation Learning}
In the first stage, we follow the DEAN objective and learn a feature representation in which the normal training data are mapped to a constant target score. To keep this from being trivial, we choose a constant of $\frac{1}{2}$. Thus, the loss we minimize is:
\begin{equation}
    L_0 = \sum_{x \in X_{\mathrm{train}}} \left(f(x) - \frac{1}{2}\right)^2 ,
    \label{eqn:l0}
\end{equation}
This loss encourages the network to produce a consistent baseline score for normal instances without considering the test data.

\paragraph{Stage 1: Transductive Fine-Tuning}
In the second stage, the model is fine-tuned using both $X_{\mathrm{train}}$ and $X_{\mathrm{test}}$. The key idea is to push the scores of test samples towards $1$, reflecting the possibility that some may be anomalous, while maintaining the $\frac{1}{2}$ target for the normal training set. The transductive objective is:
\begin{equation}
\begin{split}
    L_{1} = \frac{1}{\|X_{\mathrm{train}}\|} &\sum_{x \in X_{\mathrm{train}}} \left(f(x) - \frac{1}{2}\right)^2 \\
        + \lambda\cdot \frac{1}{\|X_{\mathrm{test}}\|} &\sum_{x \in X_{\mathrm{test}}} \left(1 - f(x)\right)^2 .
\end{split}
    \label{eqn:l1}
\end{equation}
This formulation directly maximizes the separation between the normal training distribution and the potentially contaminated test distribution. We introduce a factor $\lambda>0$ to weight the two training objectives. 
We summarize our training process in Algorithm~\ref{alg:main}.

\begin{algorithm}[H]
\caption{DOUST: Deep OUtlier Selection by Transduction}
\begin{algorithmic}[1]
\Require training data $x_{\text{train}}$ (assumed to be normal), test data $x_{\text{test}}$ (unlabeled)
\Require hyperparameters: ensemble count $N_{\text{ens}}=100$, training lengths $\text{Epochs}_{\text{pretrain}}=5$ and $\text{Epochs}_{\text{train}}=50$
\Ensure anomaly scores $\text{prediction}$

\State $(x_{\text{train}}, x_{\text{test}}) \gets$ Normalize($x_{\text{train}}, x_{\text{test}}$)

\For{$i = 0$ \textbf{to} $N_{\text{ens}}$}
\State Define neural network:
\State \quad Input layer $\rightarrow$ 3 Dense layers (100 units, ReLU, no bias) $\rightarrow$ Output layer (1 unit, modified sigmoid)

\State Pretrain model on $x_{\text{train}}$ using the Loss from Equation~\ref{eqn:l0} for $\text{Epochs}_{\text{pretrain}}$

\State Combine datasets:
\State \quad $X_{\text{combined}} \gets$ concat($x_{\text{train}}, x_{\text{test}}$)
\State \quad $Y_{\text{combined}} \gets$ concat(zeros(len($x_{\text{train}}$)),ones(len($x_{\text{test}}$)))
\State Shuffle $X_{\text{combined}}$ and $Y_{\text{combined}}$

\State Train model on $(X_{\text{combined}}, Y_{\text{combined}})$ using the Loss from Equation~\ref{eqn:l1} for $\text{Epochs}_{\text{train}}$

\State Predict:
\State \quad $\text{prediction}_i \gets$ model.predict($x_{\text{test}}$)
\EndFor

\State Merge Ensemble Predictions
\State $\text{prediction} \gets \frac{1}{N_{\text{ens}}}\sum_{i=0}^{N_{\text{ens}}} \text{prediction}_i$

\State \Return $\text{prediction}$
\end{algorithmic}
\label{alg:main}
\end{algorithm}

\subsection{Theoretical Separation Guarantee}
The following result formalizes, under idealized assumptions, that minimizing Equation~\ref{eqn:l1} yields a separation between the outputs on normal and abnormal samples. We present a concise \emph{sketch of proof} to clarify the intuition and the role of the assumptions.

\begin{lemma}[Separation under idealized conditions]
Assume (i) the large-sample limit so that the empirical loss in Equation~\ref{eqn:l1} is well-approximated by its expectation in Equation~\ref{eqn:l1Int}, (ii) $f:\mathbb{R}^d\to[0,1]$ is optimized pointwise, and (iii) the normal and abnormal densities have disjoint supports, i.e., $p_{\mathrm{normal}}(x)\,p_{\mathrm{abnormal}}(x)=0$ almost everywhere. Then any minimizer $f^\star$ of Equation~\ref{eqn:l1Int} satisfies
\[
f^\star(x)=
\begin{cases}
    1 \quad \text{for } x\in \mathrm{supp}(p_{\mathrm{abnormal}})\\
    \frac{\tfrac{1}{2}+\lambda(1-\nu)}{1+\lambda(1-\nu)} \quad \text{for } x\in \mathrm{supp}(p_{\mathrm{normal}})\\
\end{cases}
\]
Notice that this does not violate data leakage rules and is still unsupervised, since we do not use the test labels.
In particular, if $\lambda\le 1$, then $f^\star(x)\le \tfrac{3}{4}$ on normal points; which yields a strict output gap between normal and abnormal samples.
\label{lmm:separation}
\end{lemma}

\begin{proof}[Proof sketch]
In a contaminated dataset $(1-\nu)\,p_{\mathrm{normal}}+\nu\,p_{\mathrm{abnormal}}$ with contamination rate $\nu$, we can reformulate Equation~\ref{eqn:l1} in the large-sample assumption as
\begin{equation}
\begin{split}
    L_{1} \propto \int_{\mathbb{R}^d} dx\; 
     &p_{\mathrm{normal}}(x) \\
     \cdot &\left[\left(f(x) - \frac{1}{2}\right)^2 + \lambda\cdot(1 - \nu) \cdot \left(1 - f(x)\right)^2 \right] \\
    + &p_{\mathrm{abnormal}}(x) \cdot\lambda\cdot \nu \cdot \left(1 - f(x)\right)^2 
\end{split}
\label{eqn:l1Int}
\end{equation}

This equation is minimal, when on $\mathrm{supp}(p_{\mathrm{abnormal}})$ $f^\star(x)=1$ and on $\mathrm{supp}(p_{\mathrm{normal}})$:
\[
f^\star(x)=\frac{\tfrac{1}{2}+\lambda(1-\nu)}{1+\lambda(1-\nu)}.
\]
Define $t=\lambda(1-\nu)\in[0,1]$ with $\lambda\le 1$. The expression $\frac{\tfrac{1}{2}+t}{1+t}$ is increasing to a maximum of  $\tfrac{3}{4}$ at $t=1$. Thus $f^\star(x)\le \tfrac{3}{4}$ for normal points, while $f^\star(x)=1$ for abnormal points, establishing a separation of at least $\tfrac{1}{4}$ between normal and abnormal points in this idealized setting.
\end{proof}

\textbf{Intuition:} By pulling test samples to $1$ and training samples to $\tfrac{1}{2}$, we effectively pull normal samples both to $1$ and $\tfrac{1}{2}$, resulting in an equilibrium below $\tfrac{3}{4}$. Abnormal samples however, are only pulled towards $1$, resulting in a separation.

\subsection{Activation Function}
To smoothly map outputs towards the target scores, we use a shifted sigmoid activation:
\begin{equation}
    S^{+}(x) = S(x-1) = \frac{1}{1 + e^{1 - x}} ,
    \label{eqn:sigmoidp}
\end{equation}
This activation makes sure that $f(x)\in [0,1]$. While this could also have been achieved with a normal sigmoid function, using Equation~\ref{eqn:sigmoidp} makes sure that the network can not simply learn to trivially output $0$ before the activations in the pretraining step (See~\cite{DEAN}).

\subsection{Hyperparameters and Implementation Details}
We implement \textsc{Doust} in TensorFlow/Keras~\cite{tensorflow, keras}, using a fully connected network with three hidden layers of 100 units each, ReLU activations, no bias terms, and a single-unit output with a shifted sigmoid activation (Equation~\ref{eqn:sigmoidp}). Stage~0 pretraining uses the loss function $L_0$ shown in Equation~\ref{eqn:l0}. Stage~1 uses the loss function $L_{1}$ shown in Equation~\ref{eqn:l1} with weighting factor $\lambda=1$ for separating training and test contributions.  

Each model is trained with the Adam optimizer at a learning rate of $0.001$ and a batch size of $100$. Stage~0 runs for $\text{Epochs}_{\text{pretrain}}=5$ epochs and Stage~1 for $\text{Epochs}_{\text{train}}=50$ epochs.

Similar to the original DEAN method, we train each result $100$ times and combine the resulting model predictions into an ensemble $\text{prediction}=\operatorname{average}_i(\text{prediction}_i)$

Alternative loss functions and different choices of these hyperparameters are provided in Section~\ref{sec:ablation}.

\section{Evaluation}\label{sec:eval}

\subsection{Experimental Setup}
We evaluate \textsc{Doust} using the $121$ datasets proposed by ADBench~\cite{surveyzhao}, which includes a diverse collection of datasets spanning multiple domains (like medical data or astrophysics), modalities (like tabular, text and image-based), as well as different numbers of features and samples. 

Our evaluation compares \textsc{Doust} against the $21$ competitors listed in Section~\ref{sec:rw:od} and thus includes both well-established shallow methods (e.g., $k$-nearest neighbors (KNN), Isolation Forest, One-Class SVM, \ldots) and state-of-the-art deep learning models (e.g., DEAN, DAGMM, DTE, \ldots). All methods are run with recommended hyperparameter settings from their respective publications or implementations. Although slightly better hyperparameters may exist, their effect would likely be small compared to the improvements reported here~\cite{myhyperparam}.

Averaged over all ADBench datasets, training a single DOUST model takes $3\text{ seconds}$ for stage 0 pretraining, $27\text{ seconds}$ for stage 1 training, and $1\text{ second}$ for anomaly score prediction.

\subsection{Overall Performance}
\begin{figure}[htbp]
    \centerline{\includegraphics[width=1.0\linewidth]{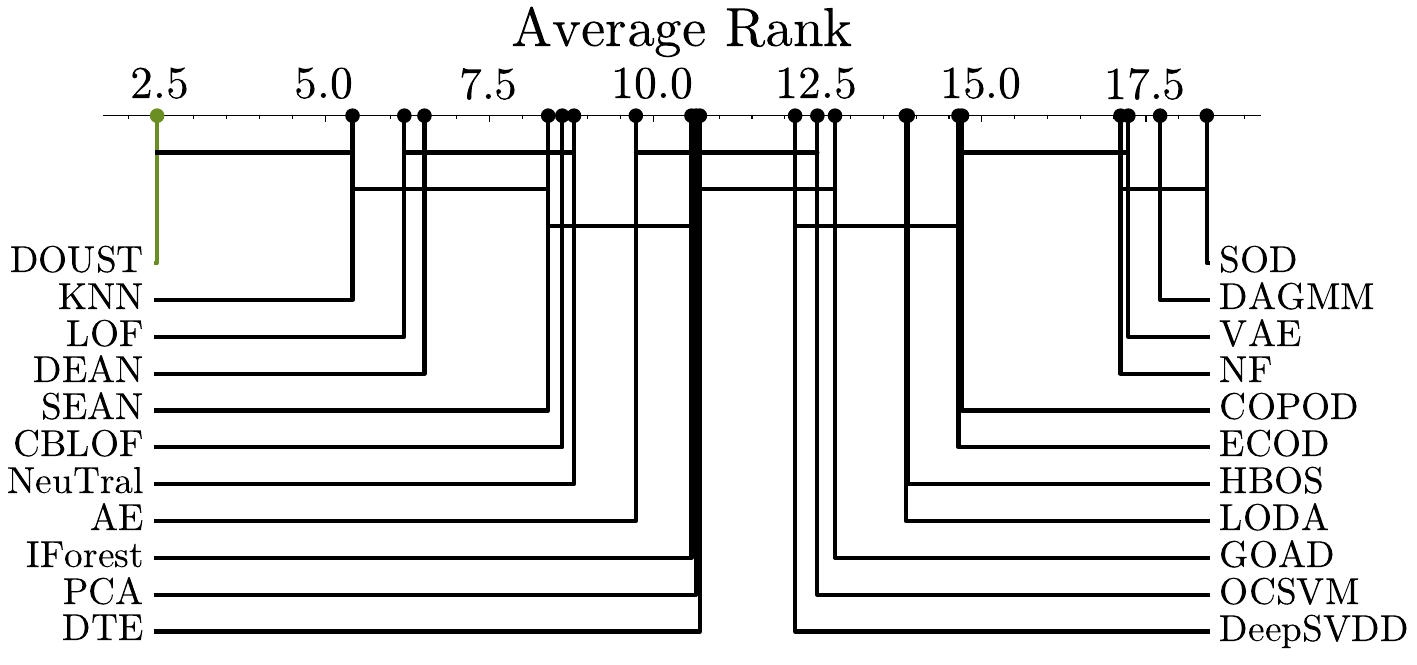}}
    \caption{Critical difference (CD) plot showing that \textsc{Doust} outperforms $21$ competitors OD methods on the ADBench benchmark. Lower ranks indicate better performance; horizontal bars show groups not significantly different per a Wilcoxon test~\cite{wilcoxon} with Bonferroni-Holm correction~\cite{correction}.}
    \label{fig:cd}
\end{figure}

Across all datasets, \textsc{Doust} achieves an average ROC-AUC of \textbf{$89.01\%$}. This is a substantial improvement over the closest competitor, KNN, which achieves $79.67\%$, corresponding to an absolute gain of nearly \textbf{$10\%$}. Figure~\ref{fig:cd} summarizes the rankings of all methods in a critical difference plot, showing that \textsc{Doust} consistently occupies the top position with statistically significant differences compared to $20/21$ competitors. We also state the individual performances in Table~\ref{tab:perf}. However, because of the large number of competitors and datasets, we only show the $8$ strongest competitors here and summarize all $121$ datasets into groups. ADBench contains multiple variations of the same datasets, which we average. The remaining datasets are grouped by their application (e.g., "Finance" and "Medical") and also averaged. DOUST outperforms each competitor in each of these groups.

\begin{table*}[]
    \centering
\begin{tabular}{llllllllll}
\hline
 Dataset           & DTE       & IForest   & CBLOF     & NeuTral   & SEAN      & DEAN      & LOF       & KNN       & \textsc{DOUST}              \\
\hline
 Medical (13 datasets)      & $67.98\%$ & $84.94\%$ & $78.96\%$ & $73.00\%$ & $79.71\%$ & $82.06\%$ & $80.81\%$ & $80.49\%$ & $\mathbf{92.09\%}$ \\
 Physical (11 datasets)     & $61.43\%$ & $76.40\%$ & $82.45\%$ & $81.13\%$ & $75.76\%$ & $78.23\%$ & $81.23\%$ & $84.55\%$ & $\mathbf{91.02\%}$ \\
 Web (9 datasets)           & $71.28\%$ & $72.40\%$ & $72.09\%$ & $73.15\%$ & $76.80\%$ & $78.74\%$ & $78.03\%$ & $76.99\%$ & $\mathbf{94.43\%}$ \\
 Image (8 datasets)         & $74.50\%$ & $80.10\%$ & $81.33\%$ & $73.84\%$ & $78.45\%$ & $81.50\%$ & $86.52\%$ & $88.59\%$ & $\mathbf{95.66\%}$ \\
 Remote (5 datasets)        & $52.61\%$ & $74.61\%$ & $73.51\%$ & $88.38\%$ & $71.46\%$ & $75.99\%$ & $89.54\%$ & $89.17\%$ & $\mathbf{97.60\%}$ \\
 Finance (4 datasets)       & $82.49\%$ & $82.24\%$ & $80.50\%$ & $65.94\%$ & $74.92\%$ & $82.57\%$ & $71.79\%$ & $84.73\%$ & $\mathbf{94.16\%}$ \\\hline
 20\_news (6 datasets)       & $49.71\%$ & $51.52\%$ & $54.71\%$ & $62.53\%$ & $61.48\%$ & $61.57\%$ & $58.20\%$ & $57.85\%$ & $\mathbf{83.03\%}$ \\
 MNIST-C (16 datasets)      & $83.59\%$ & $77.06\%$ & $82.47\%$ & $82.70\%$ & $84.48\%$ & $88.86\%$ & $91.66\%$ & $88.77\%$ & $\mathbf{95.08\%}$ \\
 FashionMNIST (10 datasets) & $90.75\%$ & $84.02\%$ & $89.61\%$ & $88.88\%$ & $90.40\%$ & $93.23\%$ & $92.16\%$ & $92.52\%$ & $\mathbf{95.36\%}$ \\
 CIFAR10 (10 datasets)      & $68.88\%$ & $63.63\%$ & $67.67\%$ & $69.72\%$ & $67.95\%$ & $72.74\%$ & $72.27\%$ & $68.28\%$ & $\mathbf{77.77\%}$ \\
 MVTec-AD (15 datasets)     & $74.19\%$ & $76.88\%$ & $77.52\%$ & $77.87\%$ & $79.88\%$ & $74.55\%$ & $77.08\%$ & $79.53\%$ & $\mathbf{81.74\%}$ \\
 agnews (4 datasets)        & $60.53\%$ & $60.10\%$ & $61.46\%$ & $64.57\%$ & $69.46\%$ & $67.18\%$ & $74.89\%$ & $67.19\%$ & $\mathbf{95.81\%}$ \\
 SVHN (10 datasets)         & $61.76\%$ & $58.57\%$ & $61.12\%$ & $62.72\%$ & $62.63\%$ & $68.71\%$ & $66.18\%$ & $64.03\%$ & $\mathbf{73.17\%}$ 
\\\hline
 Average           & $71.03\%$ & $73.86\%$ & $75.73\%$ & $75.49\%$ & $76.46\%$ & $78.70\%$ & $79.76\%$ & $79.67\%$ & $\mathbf{89.01\%}$ \\
\hline\\
\end{tabular}\vspace{-0.7em}
    \caption{Individual performance of \textsc{DOUST} compared to its strongest competitors for various groups of ADBench datasets.}
    \label{tab:perf}
\end{table*}

\subsection{Per-Dataset Superiority}
\begin{figure}[htbp]
    \centerline{\includegraphics[width=1.0\linewidth]{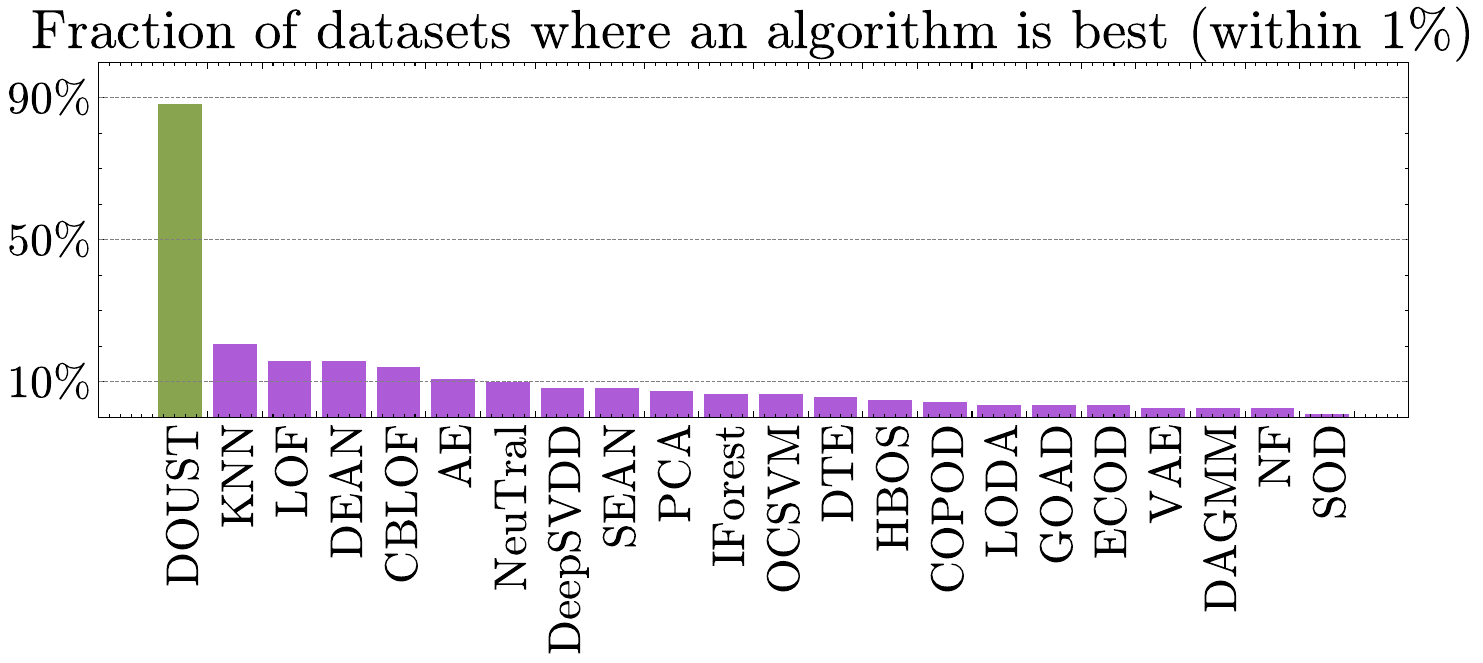}}
    \caption{Percentage of datasets on which each method achieves the highest ROC-AUC (within a $1\%$ margin). \textsc{Doust} outperforms all competitors by a wide margin, being one of the best choices on $88.43\%$ of datasets.}
    \label{fig:highest}
\end{figure}

In addition to the average performance, we also measure the frequency with which each method attains one of the highest ROC-AUC scores on a dataset. We define a method to be a good choice on a dataset if it achieves an ROC-AUC within $1\%$ of the best performing algorithm. As shown in Figure~\ref{fig:highest}, \textsc{Doust} is a good choice to use on $88\%$ of all datasets. This is more than four times as often as the next best method (KNN), which achieves this status on only $21\%$ of datasets. This dominance underscores the usefulness of our approach across a wide variety of data modalities and anomaly characteristics.

\section{Transduction and Rare Anomalies}

The results in Section~\ref{sec:eval} demonstrate that incorporating unlabeled test data can substantially improve OD performance. However, as we will show, transductive approaches inherently depend on the statistical properties of the test set, most notably its contamination rate $\nu$, i.e., the fraction of unlabeled anomalous instances in $X_{\mathrm{test}}$. 

\subsection{Dependence on Contamination Rate}
In our earlier experiments, we fixed the contamination rate $\nu = \frac{1}{2}$ for evaluation, as this setting minimizes the statistical uncertainty in performance estimation~\cite{bootstrap}. While useful for benchmarking, such a high contamination rate is unrealistic in most real-world scenarios where anomalies are usually rare.

Yet in transductive OD, the learning signal comes from differences between the training and test distributions, which become harder to detect when anomalies are rare. This tension has received surprisingly little attention in prior work. For example, \cite{lemanTransductive} only considers datasets with roughly $10\%$ anomalies and more than $2000$ samples, conditions that (as we will show later) ensure abundant anomalous examples.

\subsection{Empirical Analysis of Rare Anomalies}
\begin{figure}[htbp]
    \centerline{\includegraphics[width=1.0\linewidth]{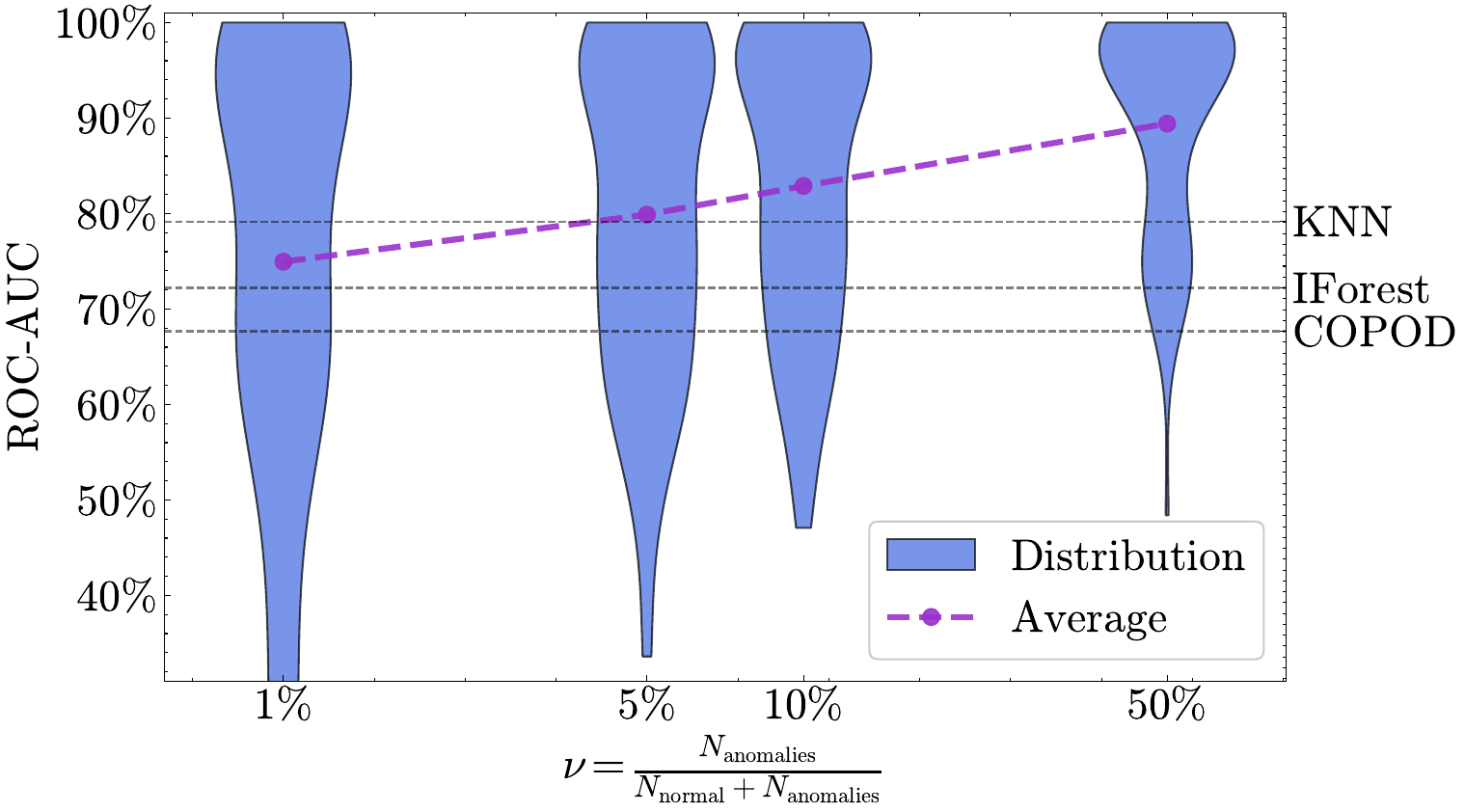}}
    \caption{Average ROC-AUC of \textsc{Doust} across varying contamination rates $\nu$. Lower contamination rates lead to decreased average performance, primarily due to occasional severe underperformance on some datasets. At very low values of $\nu$, some inductive methods outperform \textsc{Doust}.}
    \label{fig:falloff}
\end{figure}

To analyze this phenomenon, we subsample our benchmark datasets to create a performance evaluation with contamination rates $\nu \in \{1\%, 5\%, 10\%, 50\%\}$ in Figure~\ref{fig:falloff}. To do this, we restrict our analysis in this chapter to datasets with sufficient points to permit such subsampling ($92/121$ of datasets in ADBench allow this). On these datasets, we find that the average ROC-AUC of \textsc{Doust} decreases as the contamination rate $\nu$ falls, and that $\nu \geq 5\%$ is, on average, required to outperform our strongest competitor (KNN). This constitutes a significant weakness of transductive anomaly detection: the method’s advantage diminishes in precisely the low-contamination regime most common in real-world OD.

Further investigation hints at an association with the test set size. Among datasets where \textsc{Doust} achieves the best result (at $\nu=50\%$), the average $|X_{\mathrm{test}}|$ is $3125$, whereas for datasets where it does not, the average size is only $217$. This led us to hypothesize that the observed weakness arises from an insufficiently large $X_{\mathrm{test}}$. 

\subsection{Link to Detectability Theory}
Matching our intuition, a recent paper~\cite{nu} shows that detecting whether a dataset is contaminated requires either sufficiently frequent anomalies or sufficiently large datasets. This is relevant to our problem, since the optimizer in \textsc{Doust} must decide which spurious signal from the training distribution to amplify. If the contamination signal is weak, the model risks optimizing for irrelevant variations (e.g., noise) instead of the anomalies we search for.

This interpretation aligns with Figure~\ref{fig:falloff}: While the modal ROC-AUC remains stable across contamination rates, the mean drops because of a long tail of very low scores (often below the $50\%$ border of a random classifier), indicating cases where the model amplified the ``wrong'' spurious signal.

Interestingly, \cite{nu} provides a quantitative condition for detectability:
\begin{equation}
    |X_{\mathrm{test}}| \geq \frac{\alpha}{\nu^2} ,
    \label{eqn:detectability}
\end{equation}
with some constant $\alpha>1$. This would imply that for large enough test sets, even very rare anomaly signals can be reliably exploited.

\subsection{Validation of the Detectability Condition}
\begin{figure}[htbp]
    \centerline{\includegraphics[width=1.0\linewidth]{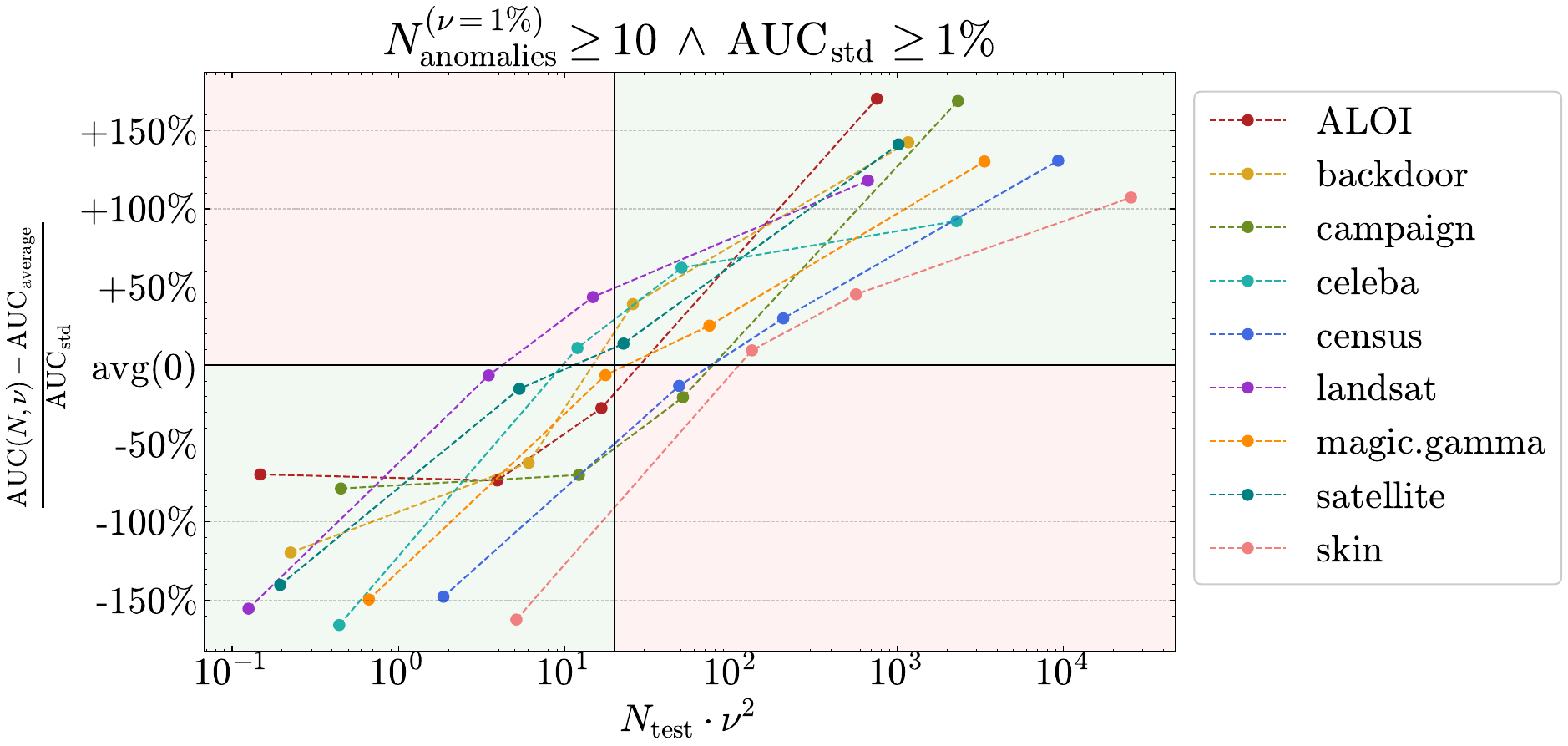}}
    \caption{Relationship between $|X_{\mathrm{test}}| \cdot \nu^2$ and the relative ROC-AUC of the \textsc{Doust} method. The results confirm the detectability condition from \cite{nu} with $\alpha \approx 20$.}
    \label{fig:ods}
\end{figure}

We test this condition using ADBench, focusing on the most reliable individual results. To reduce statistical noise, we require that each ROC-AUC measurement is based on at least $10$ anomalies and that the variation in ROC-AUC across contamination rates is sufficiently large to be informative. Plotting ROC-AUC performance against $|X_{\mathrm{test}}| \cdot \nu^2$ (Figure~\ref{fig:ods}) yields a clear threshold behavior, consistent with Equation~\ref{eqn:detectability} with $\alpha \approx 20$.

\begin{figure}[htbp]
    \centerline{\includegraphics[width=1.0\linewidth]{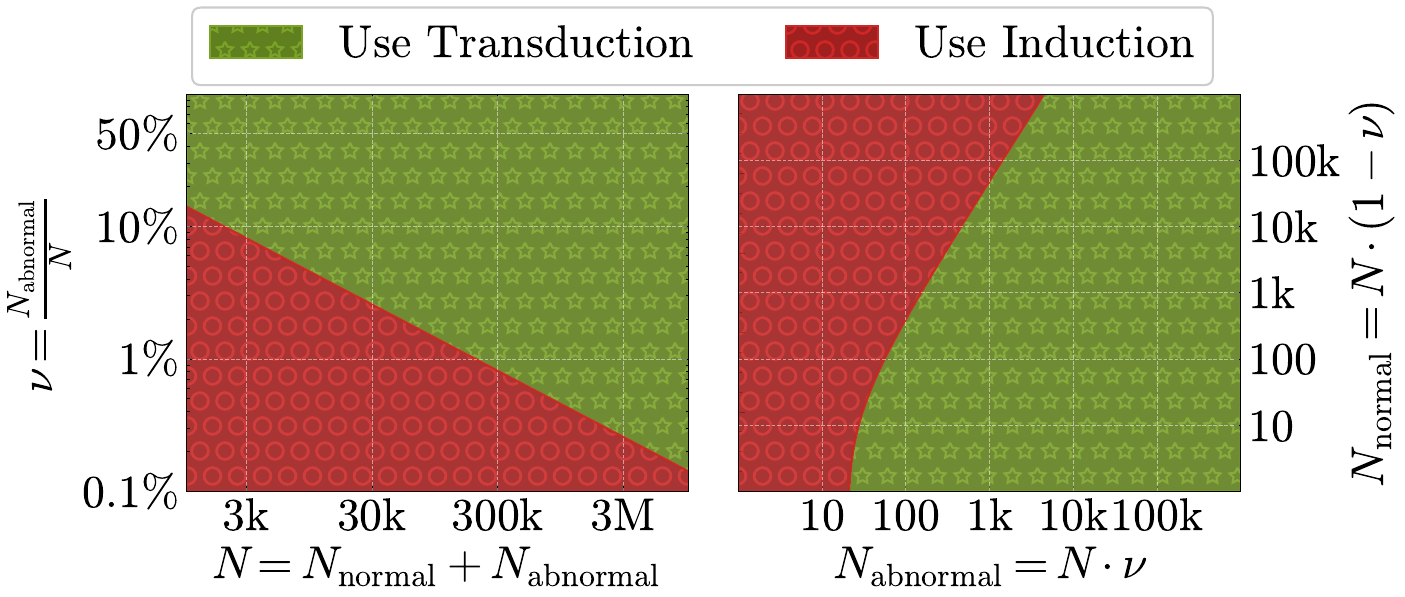}}
    \caption{Region in which using a transductive method like \textsc{Doust} improves upon inductive methods as per Equation~\ref{eqn:detectability} ($\alpha=20$). On the left, the area is shown dependent on the contamination rate $\nu$ and the total number of samples in the test set $N$, while on the right, we give the same area as a function of the number of normal and abnormal samples in the test set.}
    \label{fig:where}
\end{figure}

This threshold behaviour makes \textsc{Doust} usable in practice even when anomalies are rare. 
As an example, consider a sensor that fails with probability $1\%$ and produces one measurement per second. Applying the above condition, the required number of test samples for \textsc{Doust} to function reliably is given in time:
\[
    \frac{20}{0.01^2}\cdot 1\ \mathrm{second} = 200,000 \ \mathrm{seconds} \ \approx 2.5\ \mathrm{days}
\]
Measuring less than $3$ days of data is a reasonable tradeoff for the strongly increased OD performance of our algorithm in many industrial monitoring applications. 
Similarly, we can apply \textsc{Doust} in fraud detection, resulting in a higher performance detecting those fraud patterns, that are common and thus most damaging.
An overview of regions in which \textsc{Doust} works well can be found in Figure~\ref{fig:where}.

\section{Ablation Studies}\label{sec:ablation}

To better understand the contributions of individual design choices in \textsc{Doust}, we conducted a series of ablation experiments.

\paragraph{Loss Functions}  
Although our task can be framed as a binary classification problem, we employ a unconventional loss function in Equation~\ref{eqn:l1}. This choice is motivated by the observation that some loss functions can exhibit instability on certain datasets, which consistently result in parameters taking NaN values. Interestingly, this instability is limited to only a seemingly random subset of datasets (some variations of MVTec-AD always work, and some always fail). This effect appears for common loss functions like MSE or CrossEntropy, rendering them ineffective.

\begin{figure}[htbp]
\centerline{\includegraphics[width=1.0\linewidth]{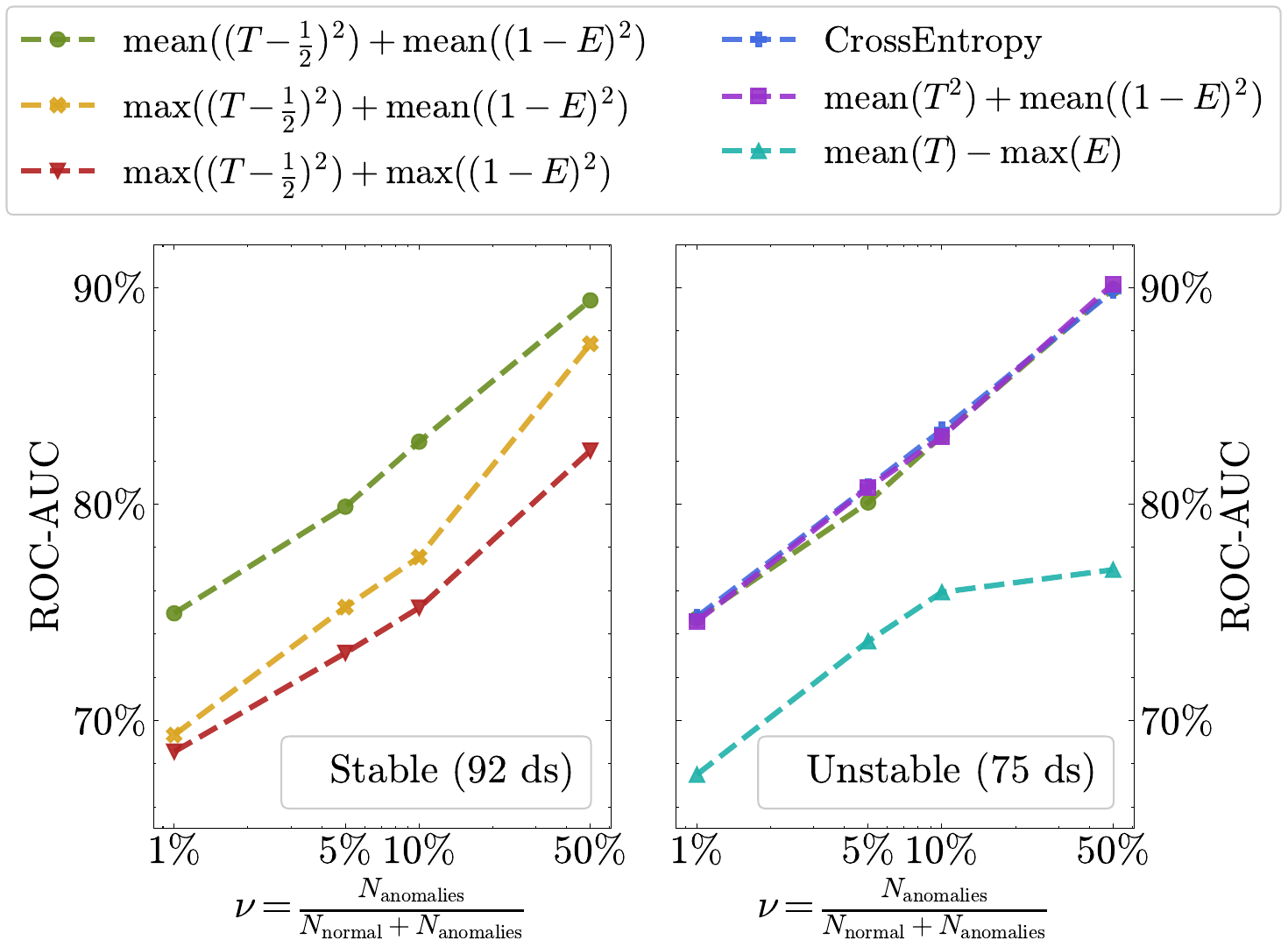}}
\caption{Comparison of different loss functions. We state the loss functions in the legend. For space reasons, $T$ represents $X_{\textrm{train}}$ and $E$ represents $X_{\textrm{test}}$. The first, green loss function is the one used in previous sections of this paper. It achieves the highest overall performance observed in our experiments and can be applied to every dataset.}
\label{fig:potential}
\end{figure}

We summarize the performance of various loss functions in Figure~\ref{fig:potential}. For those datasets, where some datasets consistently result in NaN-valued parameters, we only compare the loss functions on the $75/92$ datasets where every loss function worked. Overall, $L_1$ (green) reaches the highest performance. And while some variations, like a cross-entropy loss, would also reach an equal performance, $L_1$ is the only loss function that reaches this performance and converges without NaN values on some dataset.

\begin{figure}[htbp]
\centerline{\includegraphics[width=1.0\linewidth]{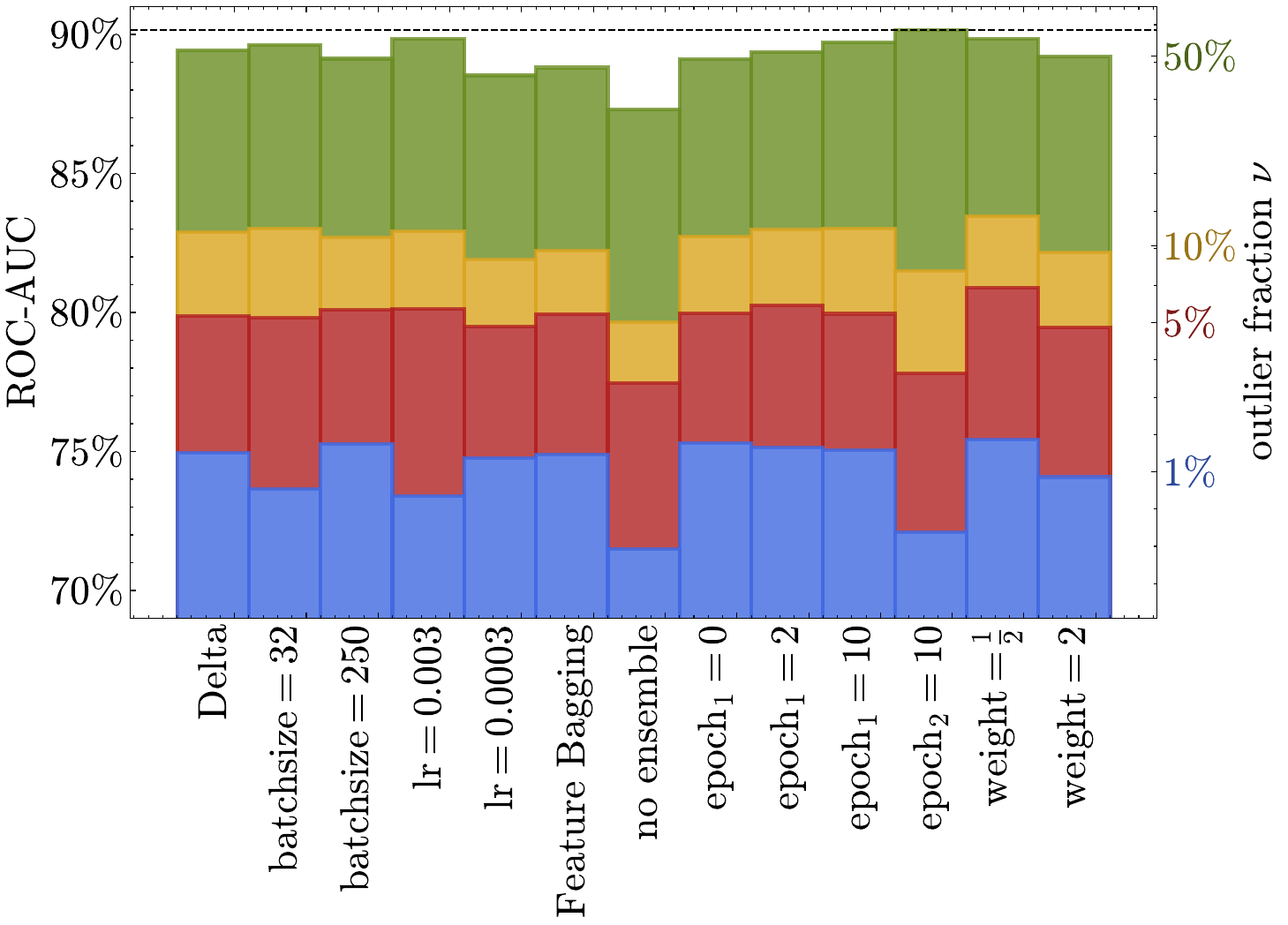}}
\caption{Hyperparameter sensitivity analysis. While some settings offer marginal improvements, the effect size is small and strongly dependent on the anomaly fraction.}
\label{fig:ablation}
\end{figure}

\paragraph{Hyperparameter Sensitivity}  
We further examined the impact of key hyperparameters. As shown in Figure~\ref{fig:ablation}, slightly higher learning rates and shorter second-stage training can yield marginal improvements. Nonetheless, these gains are highly dependent on the anomaly fraction and are generally negligible compared to the improvements achieved by the Delta loss itself. This suggests that \textsc{Doust} is robust to most hyperparameter choices, reducing the tuning burden in practice. Two interesting observations are given by the weighting factor $\lambda$. A $\lambda<1$ seems to improve upon our choice of $\lambda=1$.
Further, it seems that while not using an ensemble reduces the average performance to $87.19\%$, this difference is small ($<2\%$) and not using an ensemble decreases the runtime by a factor $100$, implying a reasonable tradeoff between speed and accuracy, which we leave for further studies.

\section{Conclusion}

In this paper, we have introduced \textsc{Doust}, the first end-to-end transductive deep learning framework for OD, and the first to explicitly leverage transduction to enhance an existing OD method. Our experiments demonstrate that, under favorable conditions, the resulting performance improvements can be substantial, outperforming all baselines by wide margins.

In addition to the strength of our method, we are also the first to systematically analyze the drawbacks of transduction in the presence of rare classes, as is typical in anomaly detection. Our results reveal that transductive methods can underperform when anomalies are extremely scarce; however, this limitation vanishes for sufficiently large datasets, following the relationship $N > \frac{\alpha}{\nu^2}$ between dataset size $N$ and contamination rate $\nu$, with a constant $\alpha\approx20$. Overall, our findings reveal an intriguing trade-off: when data is abundant, \textsc{Doust} detects anomalies with extremely high precision, and when data is scarce, inductive baselines may be preferable. 

We believe that our work represents only an initial exploration of transductive anomaly detection. Future research should focus on developing improved loss functions, further understanding the detection limits in low-contamination settings, and exploring other applications of our paradigm. For instance, our loss formulation could be directly applied to model selection in OD, extending beyond the parameter-specific approach of \cite{lemanTransductive} to enable comparison across arbitrary hyperparameter or model choices.


    


\bibliographystyle{IEEEtran}
\bibliography{02_literature,indiv,literature-review,main,new,publications,randomlymissing,theses,ai}

\end{document}